\pdfoutput=1

\documentclass[11pt]{article}

\usepackage[final]{acl}

\usepackage{times}
\usepackage{latexsym}

\usepackage[T1]{fontenc}

\usepackage[utf8]{inputenc}

\usepackage{microtype}

\usepackage{inconsolata}

\usepackage{graphicx}

\usepackage{subcaption}
\usepackage{comment}
\usepackage{lipsum}
\usepackage{multirow}
\usepackage{makecell}
\usepackage[noend,linesnumbered,ruled,vlined]{algorithm2e}
\usepackage{MnSymbol}

\newcommand{\llm}{\textsc{llm}}
\newcommand{\ie}{\textit{i.e.}}
\newcommand{\eg}{\textit{e.g.}}
\newcommand{\mathfh}{M\textsc{ath}$500$}
\newcommand{\aimetfr}{A\textsc{ime}$24$}
\newcommand{\aimetfv}{A\textsc{ime}$25$}
\newcommand{\aime}{A\textsc{ime}}
\newcommand{\granitesmall}
{\textsc{g}ranite$3.3$-$2$\textsc{b}-\textsc{i}nstruct}
\newcommand{\granitebig}
{\textsc{g}ranite$3.3$-$8$\textsc{b}-\textsc{i}nstruct}
\newcommand{\qwensmall}
{\textsc{q}wen$2.5$-\textsc{m}ath-$1.5$\textsc{b}-\textsc{i}nstruct}
\newcommand{\qwenbig}{\textsc{q}wen$2.5$-\textsc{m}ath-$7$\textsc{b}-\textsc{i}nstruct}
\newcommand{\phifour}{\textsc{p}hi$4$}
\newcommand{\chot}{\textsc{c}o\textsc{t}}

\title{Confidence-Weighted Token Set Cover for\\Early Hypothesis Pruning in Self-Consistency}

\author{Md Arafat Sultan$^\diamond$\quad\quad~Ramón Fernandez Astudillo\\IBM Research AI\\ \texttt{$^\diamond$arafat.sultan@ibm.com}\\}

\begin{document}
\maketitle

\begin{abstract}
Despite its simplicity and efficacy, the high token expenditure of self-consistency can limit its practical utility.
Here we investigate if self-consistency can be made more token-efficient for long chain-of-thought reasoning tasks, \textit{while preserving its parallelism}, through early hypothesis pruning.
Concretely, we generate all solutions in parallel, but periodically prune intermediate hypotheses that are deemed unnecessary based on two lightweight indicators: (\textit{a}) the model's own confidence in individual hypotheses, and (\textit{b}) lexical coverage of all current hypotheses by candidate subsets that are under consideration for continued retention.
We design a fast weighted set cover algorithm that utilizes the two indicators; our evaluation of five \llm{}s on three math benchmarks shows that this method can improve token efficiency for all models, by $10-35\%$ in many cases.
\end{abstract}

\section{Self-Consistency and Efficiency}
\label{section:introduction}

Self-consistency, also referred to as majority voting, solves a given problem by sampling multiple solutions and selecting the most frequent answer in those solutions as the final output \cite{wang2023selfconsistency}.
A highly effective test-time scaling strategy for weak and strong models alike on a variety of reasoning tasks \cite{touvron2023llama,deepseekai2025deepseekr1,mistralai2025magistral}, it is also in many ways a lightweight and low-cost method -- it does not need a separate scoring model, unlike methods such as best-of-$N$ (\textsc{b}o\textsc{n}) sampling or beam search, for example. 
Yet, generating a large number of samples can itself be computationally expensive, especially with modern large language models (\llm{}s).

Existing approaches to sample-efficient majority voting are predominantly recurrent in nature: given a sample budget $N$, a small batch of $M<<N$ complete solutions is generated at a time; the process is repeated until strong consensus is observed among already generated samples, or in the worst case, the budget has been fully spent \cite{aggarwal2023lets,li2024escape}.
Later work has incorporated utilities of the generated samples into the process, sourced either from the generator's own confidence scores \cite{wang2024soft,huang2025efficient} or from an external scorer, such as a feature-based classifier \cite{wan2025reasoning} or a reward model \cite{astudillo2025optimal}. 
Despite improving sample efficiency, the sequentiality of these methods remains their clear drawback, as turnaround time increases linearly with the number of batches.

Given today's \textsc{gpu}-powered, highly parallelized \llm{} compute environments, here we are interested in token-efficient self-consistency that better retains parallelism.
Our specific context is that of reasoning tasks such as math problem solving, where solutions take the form of a relatively long chain of thought (\chot{}) followed by a short final answer.
Concretely, we consider an approach where all solutions are still generated in parallel as in standard self-consistency, but hypotheses -- sequences of generated tokens -- unlikely to add to the predictive power of the cohort are pruned early.
Similar strategies have been used to prune low-scoring hypotheses for efficient \textsc{b}o\textsc{n} sampling, \eg{}, by \citet{sun2024fast} and \citet{wang2025samplingefficient}.
Self-consistency poses different challenges than \textsc{b}o\textsc{n}, however, as an entire subset of hypotheses must be maintained through the end that are not only of high quality individually, but also diverse \cite{lau2024dipper,wang2025diversified} and sufficiently exploratory of the space of solutions collectively.

The key challenge in our approach lies in identifying the prunable hypotheses: Given a set $Y_t$ of parallel hypotheses for an input $x$ at timestep $t$, what subset $Y_t^* \subset Y_t$ should be retained for further expansion so that the rest 
can be safely pruned?
We posit that a combination of two complementary properties can make strong $Y_t^*$ candidates: (\textit{i})~Individual hypotheses in $Y_t^*$ are of high quality, and (\textit{ii})~$Y_t^*$ retains the full diversity of ``thoughts'' verbalized across all hypotheses in $Y_t$.
Using computationally inexpensive proxies to quantify the underlying attributes -- the \llm{}'s own token-level probabilities for individual hypothesis quality and the lexical coverage of $Y_t$ by $Y_t^*$ for group diversity -- we propose an approximate confidence-weighted token set cover algorithm for subset selection (\S{\ref{section:method}}) that allows us to examine our proposition.

Experimental results (\S{\ref{section:experiments}}) show that the proposed method can improve token efficiency for five different \llm{}s in the $1.5$\textsc{b}-$14$\textsc{b} parameters range from the \textsc{q}wen$2.5$-\textsc{m}ath \cite{yang2024qwen2.5}, \textsc{g}ranite$3.3$\footnote{\scriptsize \url{https://huggingface.co/ibm-granite/granite-3.3-8b-instruct}} and \phifour{} \cite{abdin2024phi4} families on multiple math benchmarks, often in the $10-35\%$ range for $32-64$ samples.
We also report ablation results validating the individual utility of hypothesis quality and diversity, and uncover model attributes that dictate the efficacy of the method by analyzing the outcome of its individual steps.

\section{Method}
\label{section:method}

\SetAlCapFnt{\small}

\begin{algorithm}[t]
\SetAlgoLined
\DontPrintSemicolon
\small
\SetKwInOut{Input}{Input}
\SetKwInOut{Output}{Output}

\Input{Universe $U$ of individual items, list of sets of items $S=[S_i]_{i=1}^N$ and their weights $[w_i]_{i=1}^N$}
\Output{Weighted set cover $C \subset S$ of $U$ (approx.)}

$PQ \leftarrow$ An empty min-priority queue\;

\For{$S_i \in S$}{
    Enqueue $S_i$ into $PQ$ with priority $\frac{w_i}{|S_i|}$\;
}

$C \leftarrow \emptyset$\quad\tcp*[h]{}\textit{the cover (set of sets from $S$)}\;
$Covered \leftarrow \emptyset$\quad\tcp*[h]{}\textit{set of items from $U$}\;

\While{$Covered \neq U$}{
    $S^* \leftarrow \text{Dequeue}(PQ)$\;
    \If{$S^* - Covered \neq \emptyset$}{
        $C \leftarrow C \cup \{S^*\}$\quad\tcp*[h]{}\textit{add $S^*$ to $C$}\;
        $Covered \leftarrow Covered \cup S^*$\quad\tcp*[h]{}\textit{new items}\;
    }
}

\Return $C$\;

\caption{\small \textsc{WeightedSetCover}}
\label{algorithm:greedy-weighted-set-cover-pq}
\end{algorithm}

\begin{algorithm}[t]
\SetAlgoLined
\DontPrintSemicolon
\small
\SetKwInOut{Input}{Input}
\SetKwInOut{Output}{Output}

\Input{Question $x$, \llm{} $\theta$, sample budget $N$, step size schedule {$\breve C$}}
\Output{Answer $a$}

$t \leftarrow 0$\;
$Y_t \leftarrow [\varepsilon]_{i=1}^N$\quad\tcp*[h]{}\textit{list of $N$ empty strings}\;
\While{$Y_t$ contains incomplete hypotheses}{ \label{algorithm2:main-loop}
    $ss \leftlsquigarrow \breve C$\quad\tcp*[h]{}\textit{get step size for current iteration}\; \label{algorithm2:get-ss}
    \For{i in $1:N$}{
        \If{$y_i$ is incomplete}{
            $y_{i,t+1:t+ss} \sim P_\theta(\cdot \mid x, y_{i, \leq t},ss)$\;
        }
    }
    $t \leftarrow t + ss$\; \label{algorithm2:eog}
    \For{i in $1:N$}{ \label{algorithm2:subset-start}
        $S_i \leftarrow \texttt{unique-tokens}(y_{i,1:t})$\;
        $conf_i \leftarrow e^{\frac{1}{t}\sum_{j=1}^{t}\text{log}P_\theta(y_{i,j}|x_i,y_{i,<j})}$\; \label{algorithm2:conf}
        $w_i \leftarrow 1 - conf_i$ \label{algorithm2:weight}
    }
    $U \leftarrow \bigcup_{i=1}^N S_i$\;
    $keep \leftarrow \textsc{WeightedSetCover}(U, [S_{i}]_{i=1}^N, [w_i]_{i=1}^N)$\; \label{algorithm2:call-wsc}
    $Y_t^* \leftarrow [y_i \mid (y_i \in Y_t) \land (S_i \in keep)]$\; \label{algorithm2:subset-end}
    \label{algorithm2:subset-end}
    $N \leftarrow \texttt{num-elements}(Y^*_t)$\;
    $Y_t \leftarrow Y_t^*$
}
\For{$y_i \in Y_t^*$}{ \label{algorithm2:fa-start}
    $a_i \leftarrow \texttt{extract-final-answer}(y_i)$
}
$a \leftarrow \texttt{majority-element}([a_i]_{i=1}^{N})$\; \label{algorithm2:fa-end}
\Return $a$ \label{algorithm2:return-fa}  

\caption{\small\textsc{SelfConsistencyWithPruning}}
\label{algorithm:efficient-self-consistency}

\end{algorithm}

At a high level, our proposed method operates by generating multiple parallel hypotheses in a step-by-step fashion and pruning a subset of those after every step, as shown in Algorithm~$\ref{algorithm:efficient-self-consistency}$.
The input to the algorithm is a question $x$, an \llm{} $\theta$ to answer it with, the sample budget $N$, and a schedule $\breve C$ that dictates the step size, \ie{}, the number of tokens to generate at every step.
For example, the step size can be kept fixed throughout execution or be gradually lowered to increase the frequency of pruning at later stages.
Each iteration of the algorithm (line $\ref{algorithm2:main-loop}$) consists of first generating $ss$ (step size) next tokens for each surviving incomplete hypothesis in $Y_t$, where $t$ is the current number of tokens per hypothesis (lines $\ref{algorithm2:get-ss}$--$\ref{algorithm2:eog}$), and then identifying a subset $Y^*_t \subset Y_t$ to retain for the next iteration (lines $\ref{algorithm2:subset-start}$--$\ref{algorithm2:subset-end})$.
The loop is exited when all hypotheses in $Y_t$ are complete, \ie{}, contain a final answer.
A majority vote is then performed on the final answers extracted from all solutions in $Y_t$ and the most frequent answer is returned (lines $\ref{algorithm2:fa-start}$--$\ref{algorithm2:return-fa}$).

The subset $Y^*_t$ is identified using a weighted set cover algorithm that is given three inputs: the set $U$ of all unique tokens across all current hypotheses in $Y_t$, a list $S=[S_i]_{i=1}^N$ where $S_i$ is the set of unique tokens in hypothesis $y_i \in Y_t$, and the weight $w_i$ of $S_i$ (line $\ref{algorithm2:call-wsc}$).
$w_i$ is computed as $1-conf_i$, where $conf_i \in [0, 1]$ is the model's length-normalized aggregate confidence in $y_i$ (lines $\ref{algorithm2:conf}$, $\ref{algorithm2:weight}$).
Algorithm $\ref{algorithm:greedy-weighted-set-cover-pq}$ details our fast approximate set cover implementation.
All token sets $S_i \in S$ are first inserted into a min-priority queue $PQ$ according to both their coverage $|S_i|$ of $U$ and $w_i$.
With a binary heap implementation, populating $PQ$ takes $O(N)$ time.
Token sets are then dequeued from $PQ$ one at a time and added to the cover $C$ if they contain items not in $Covered$, the set of already covered tokens.
The loop terminates as soon as $Covered=U$, running in $O(N\text{log}N)$ worst-case time.
This method essentially prefers to have hypotheses with jointly high confidence and coverage scores in $C$ provided they add to the ongoing coverage of $U$.

\begin{figure*}[ht]
  \centering
  \begin{subfigure}{.25\linewidth}
    \centering
    \includegraphics[width=\linewidth]{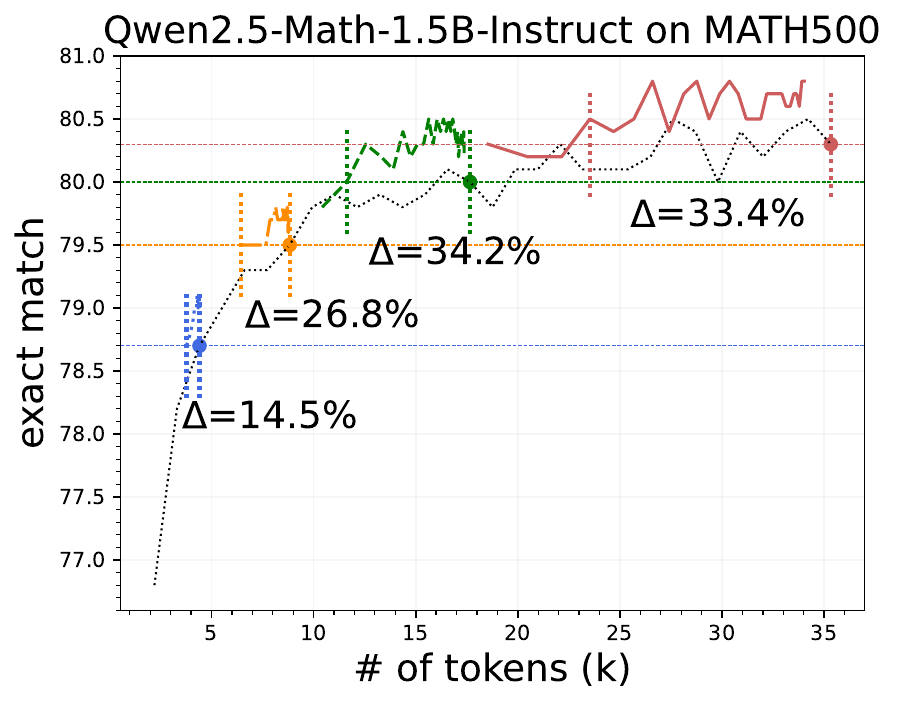}
  \end{subfigure}%
  \begin{subfigure}{.25\linewidth}
    \centering
    \includegraphics[width=\linewidth]{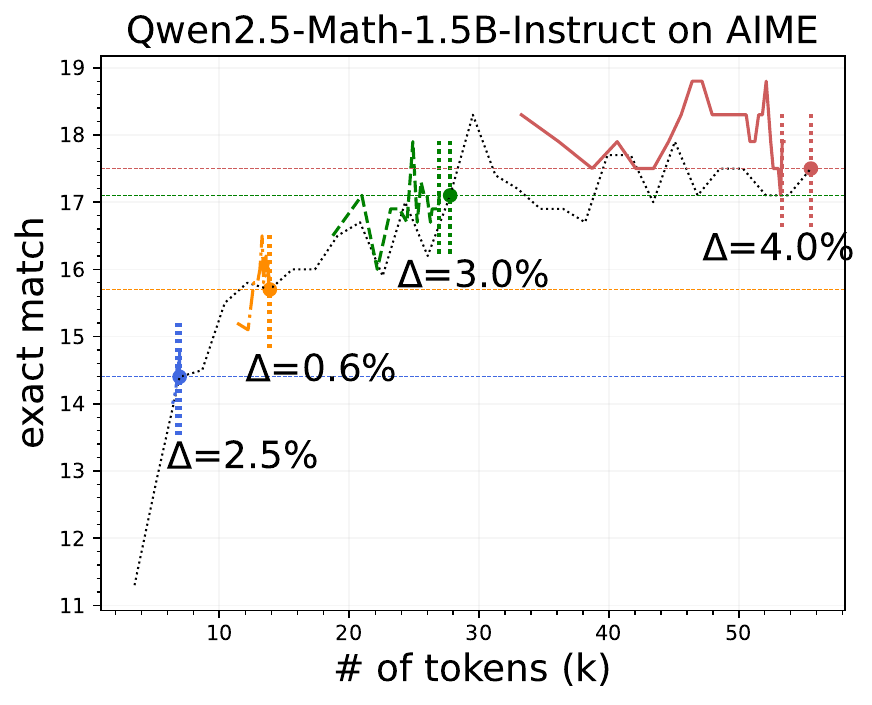}
  \end{subfigure}%
  \begin{subfigure}{.25\linewidth}
    \centering
    \includegraphics[width=\linewidth]{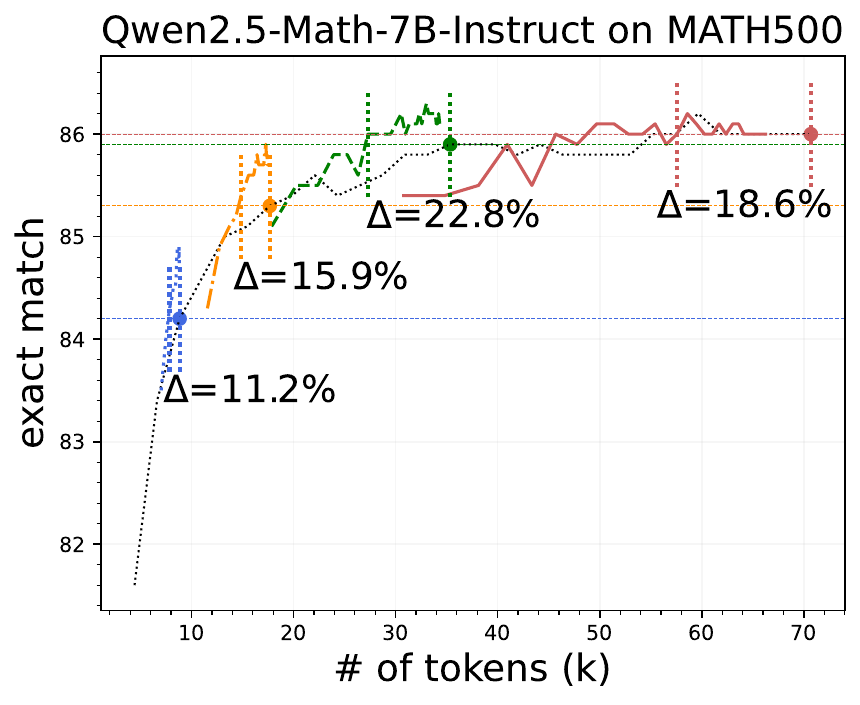}
  \end{subfigure}%
  \begin{subfigure}{.25\linewidth}
    \centering
    \includegraphics[width=\linewidth]{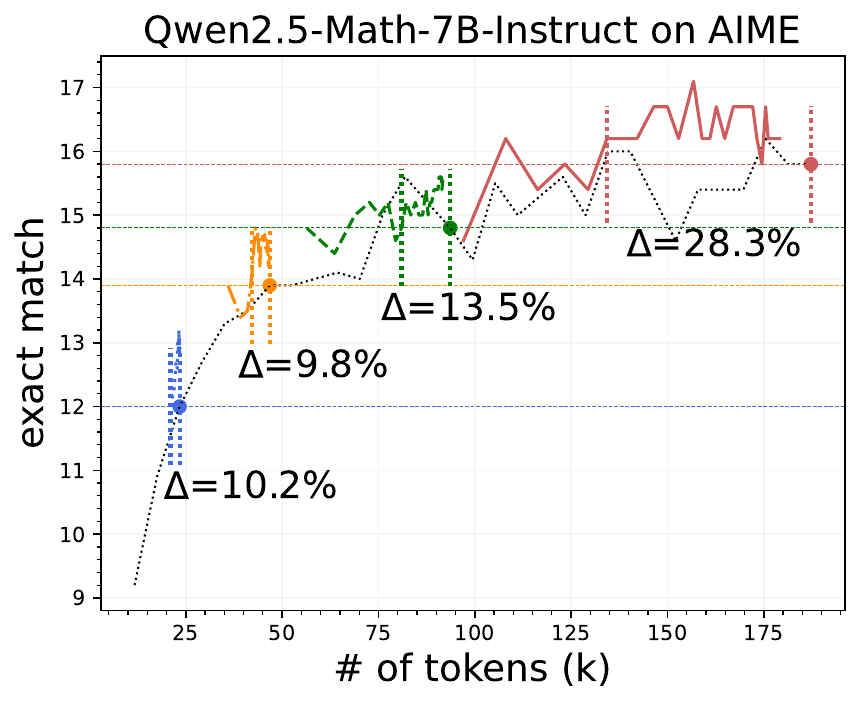}
  \end{subfigure}
  \begin{subfigure}{.25\linewidth}
    \centering
    \includegraphics[width=\linewidth]{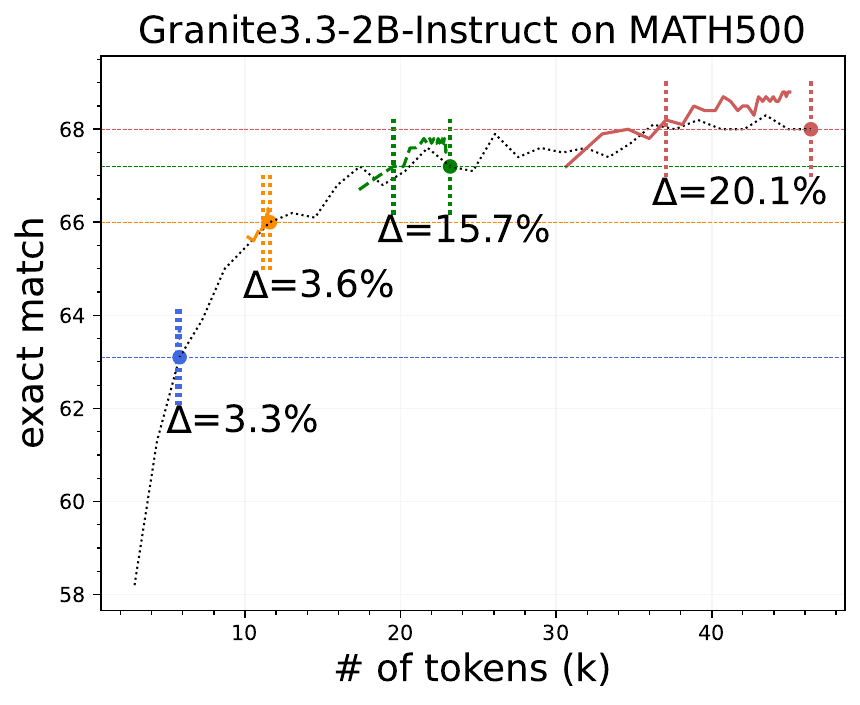}
  \end{subfigure}%
  \begin{subfigure}{.25\linewidth}
    \centering
    \includegraphics[width=\linewidth]{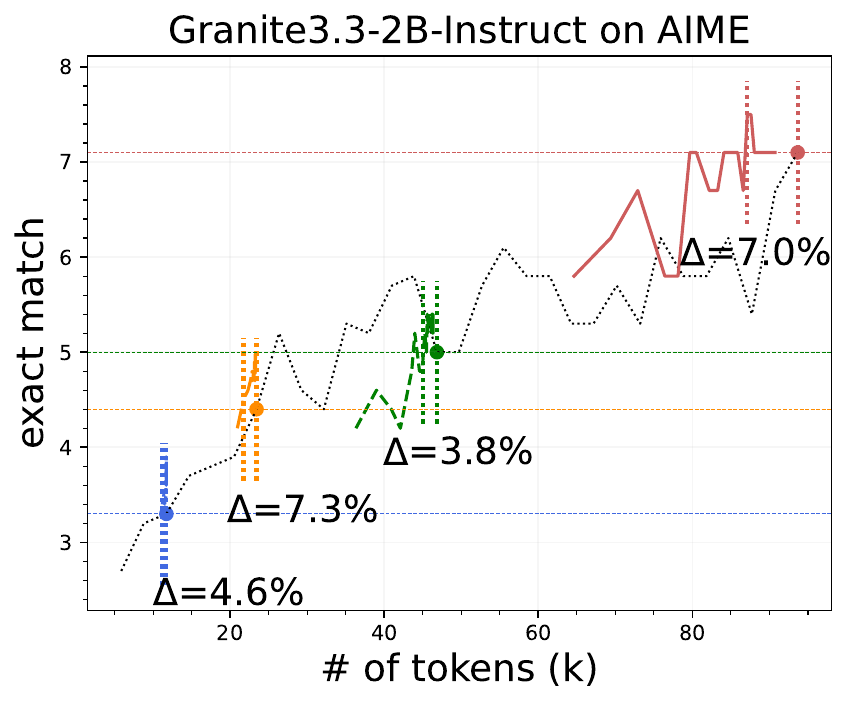}
  \end{subfigure}%
  \begin{subfigure}{.25\linewidth}
    \centering
    \includegraphics[width=\linewidth]{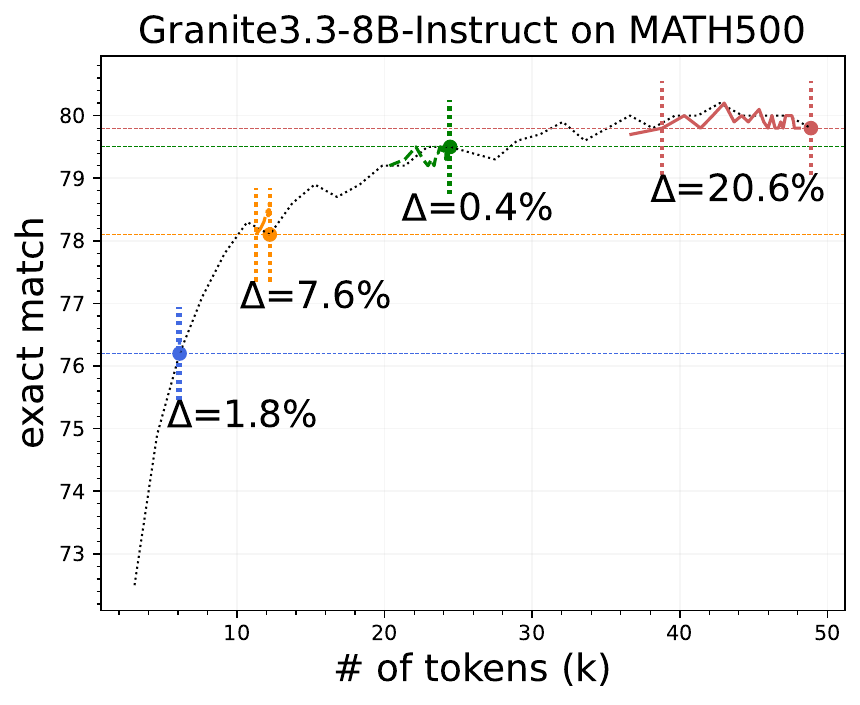}
  \end{subfigure}%
  \begin{subfigure}{.25\linewidth}
    \centering
    \includegraphics[width=\linewidth]{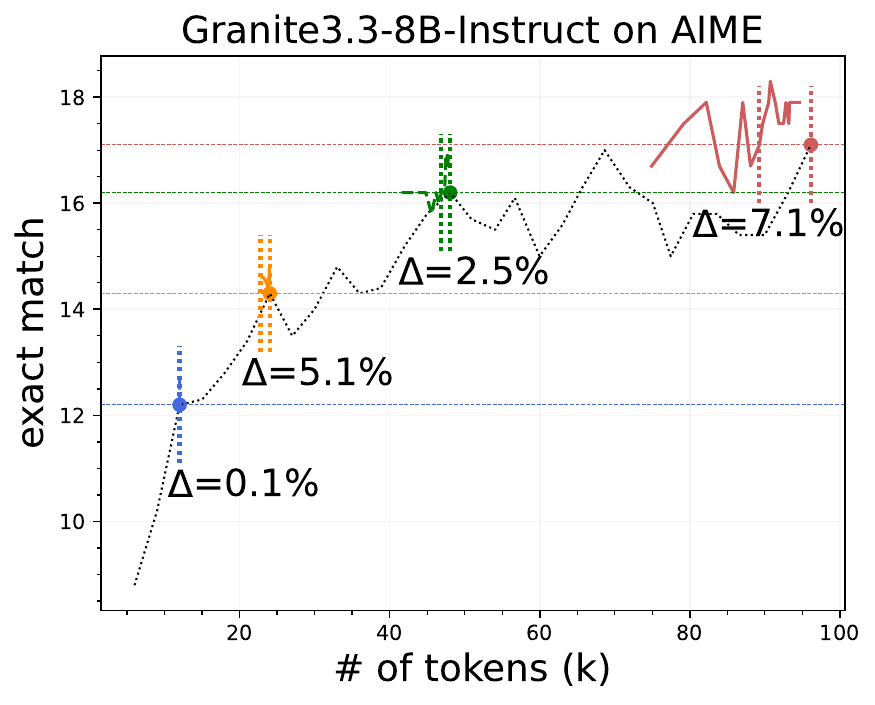}
  \end{subfigure}
  \begin{subfigure}{.25\linewidth}
    \centering
    \includegraphics[width=\linewidth]{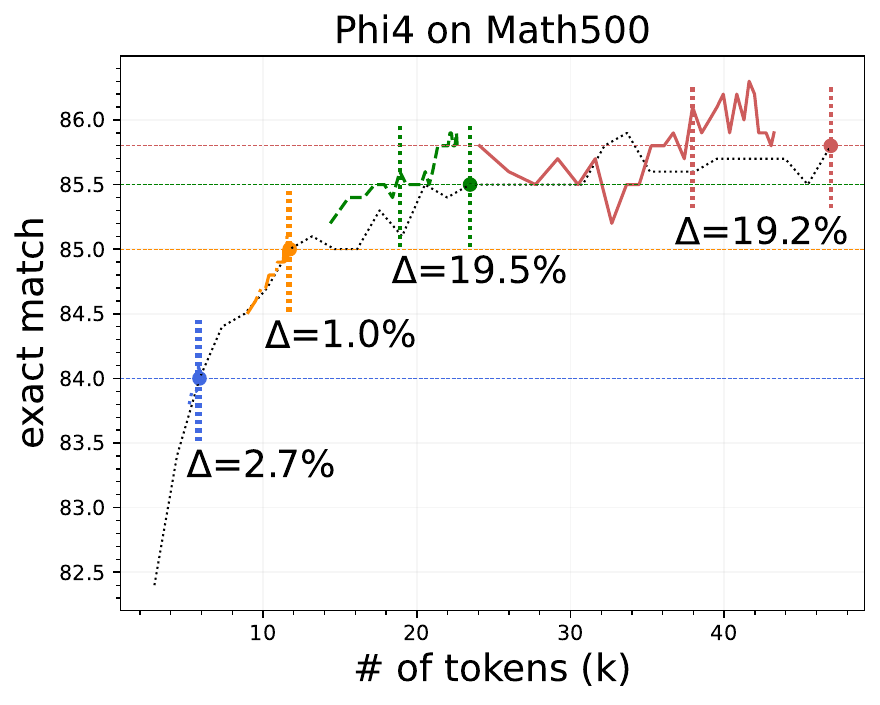}
  \end{subfigure}
  \begin{subfigure}{.45\linewidth}
    \centering
    \includegraphics[width=\linewidth]{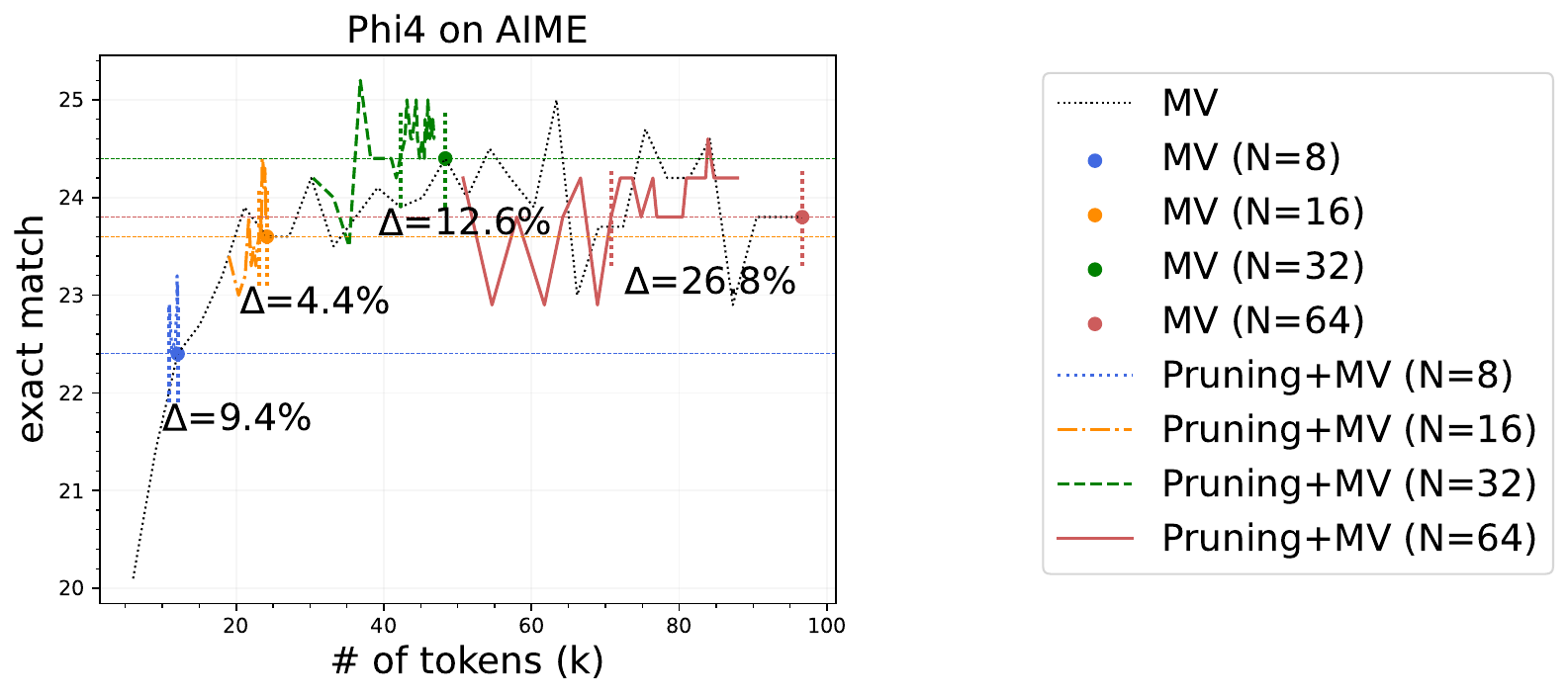}
  \end{subfigure}  
  \caption{Token savings from early pruning. The four short colored polylines in each plot correspond to hypothesis pruning at sample budgets $N=8, 16, 32, 64$. Each polyline depicts exact match (\textsc{em}) as step size (and consequently, the number of tokens generated) increases. For each, $\Delta$ represents the $\%$ of tokens saved when it passes the corresponding majority voting \textsc{em} without pruning (horizontal baseline) and subsequently never drops below it. The $\Delta$ values are also summarized in Table~\ref{table:pct-token-savings} (Appendix \ref{appendix:token-savings}). The long black dotted polyline depicts ordinary majority voting \textsc{em} at different sample budgets $N \in [4, 64]$, and crucially, stays mostly below the early pruning polylines.}
  \label{figure:token-savings-with-pruning}
\end{figure*}

\section{Experiments}
\label{section:experiments}

\paragraph{\textbf{Setup:}}\par\hspace{-.25cm}We evaluate five different \llm{}s of various sizes on three public math benchmarks of varying difficulty levels.
The \llm{}s include Alibaba's math-specific \textsc{q}wen$2.5$-\textsc{m}ath-\textsc{i}nstruct with $1.5$\textsc{b} and $7$\textsc{b} parameters, IBM's general-purpose \textsc{g}ranite$3.3$-\textsc{i}nstruct with $2$\textsc{b} and $8$\textsc{b} parameters that have strong math performance for their size and class, and Microsoft's \phifour{} with $14$\textsc{b} parameters as our largest model.
Our benchmarks are \mathfh{} \cite{hendrycks2021measuring,lightman2023let} with $500$ test problems, and \aimetfr{}\footnote{\scriptsize \url{https://huggingface.co/datasets/math-ai/aime24}} and \aimetfv{}\footnote{\scriptsize \url{https://huggingface.co/datasets/math-ai/aime25}} with $30$ problems each.
We merge the two \aime{} sets into one and report results on the merged set.
All model predictions are evaluated against ground truth answers using exact match (\textsc{em}) via a custom fork of \texttt{math-evaluation-harness}\footnote{\scriptsize \url{https://github.com/ZubinGou/math-evaluation-harness}}.

We report results for sample budgets $N$ $=$ $8$, $16$, $32$ and $64$.
For each problem, we generate a total of $256$ samples with each \llm{} (temperature $1.0$, top-$p$ $0.95$) from $16$ different seeds so that evaluation scores can be averaged over multiple runs, \eg{}, $16$ runs for $N=16$ and $4$ runs for $N=64$.
See Appendix \ref{appendix:token-counts} for length statistics.
Finally, we adopt a step size schedule $\breve C$ that starts with a pre-specified initial step size $ss$ and reduces it by half after every step as more hypotheses get pruned, until $ss$ reaches a minimum ($8$ in our experiments).
This schedule essentially applies pruning more often at later stages of inference as hypotheses get longer and more representative of their final forms.

\paragraph{Results:}\par\hspace{-.25cm}Figure \ref{figure:token-savings-with-pruning} shows our main results.
Each plot depicts the accuracy of the proposed method (\textsc{p}runing+\textsc{mv}) and of ordinary majority voting with no pruning (\textsc{mv}) for a unique model-test set pair with sample budgets $N = 8, 16, 32, 64$.
For each value of $N$, we run \textsc{p}runing+\textsc{mv} with different initial step sizes ranging from $64$ to $512$, shown as a short colored polyline in the plots.
It should be noted here that smaller step sizes generally lead to more aggressive early hypothesis pruning -- as groups of short solution prefixes tend to be more homogeneous than longer ones -- resulting in high token efficiency but low accuracy.
We use the symbol $\Delta$ and two parallel vertical dotted lines in the plots to show token savings, which is computed as the $\%$ by which \textsc{p}runing+\textsc{mv} reduces token count from \textsc{mv}.
This is determined at the point where the \textsc{p}runing+\textsc{mv} \textsc{em} score passes that of \textsc{mv} and subsequently never falls behind.
\textsc{p}runing+\textsc{mv} almost always yields token savings, by a sizable $10-35\%$ in many cases.
Savings are understandably higher at larger values of $N$, but are non-negligible $(>5\%)$ even at lower budgets in many instances.

The long black dotted polyline in each plot shows the accuracy of vanilla \textsc{mv} for different token counts, corresponding to different values of $N \in [4, 64]$.
Crucially, more often than not, and especially within the $\Delta$ range, the \textsc{p}runing$+$\textsc{mv} polylines stay above this \textsc{mv} polyline for an equal number of generated tokens, indicating that the method can be robust to the selection of step size.

\begin{table}[t]
    \centering
    \small
    \begin{tabular}{cccc}
        \textbf{Method 1} & \textbf{Method 2} & \textbf{Winner} & \textbf{Win $\%$} \\
        \Xhline{1.6\arrayrulewidth}
        \textsc{cwsc} & \textsc{sc} & \textsc{cwsc} & $82.5$ \\
        \textsc{cwsc} & \textsc{cw} & \textsc{cwsc} & $65.0$ \\
        \textsc{sc} & \textsc{cw} & \textsc{cw} & $62.5$ \\
        \Xhline{1.6\arrayrulewidth}
    \end{tabular}
    \caption{Ablation results. \textsc{cwsc}: confidence-weighted set cover (proposed); \textsc{sc}: unweighted set cover; \textsc{cw}: confidence weighting with no set cover. Both confidence weighting and set cover contribute to the overall performance of \textsc{cwsc} (rows $1, 2$), with the former playing a more important role (rows $1, 2, 3$).}
    \label{table:ablation}
\end{table}

\textbf{\textit{Ablation.}}
To gauge the individual importance of confidence weighting and token set cover for the algorithm, we ablate each and evaluate the resulting method.
For \textit{set cover only}, we assign a weight of $1$ to all hypotheses uniformly in line $\ref{algorithm2:weight}$ of Algorithm $\ref{algorithm:efficient-self-consistency}$.
For \textit{confidence weighting only}, we substitute the set $keep$ in line $\ref{algorithm2:call-wsc}$ with the same number of highest-confidence hypotheses.

Across our $5$ \llm{}s, $2$ test sets and $4$ different sample budgets, we have a total of $40$ test setups.
By first averaging over various amounts of tokens generated within each setup, and then further averaging over all $40$ setups, we compare the performance of different methods in Table \ref{table:ablation}.
Row $1$, for example, shows that the full method (\textsc{cwsc}) outperforms \textit{set cover only} (\textsc{sc}) in $82.5\%$ of all setups.
Rows $1$ and $2$ validate the need for both components in the algorithm, while all rows together point to confidence weighting as the more important one. See also Appendix \ref{appendix:compare-with-random} for random pruning results.

\begin{figure}[t]
  \centering
  \begin{subfigure}{.75\linewidth}
    \centering
    \includegraphics[width=\linewidth]{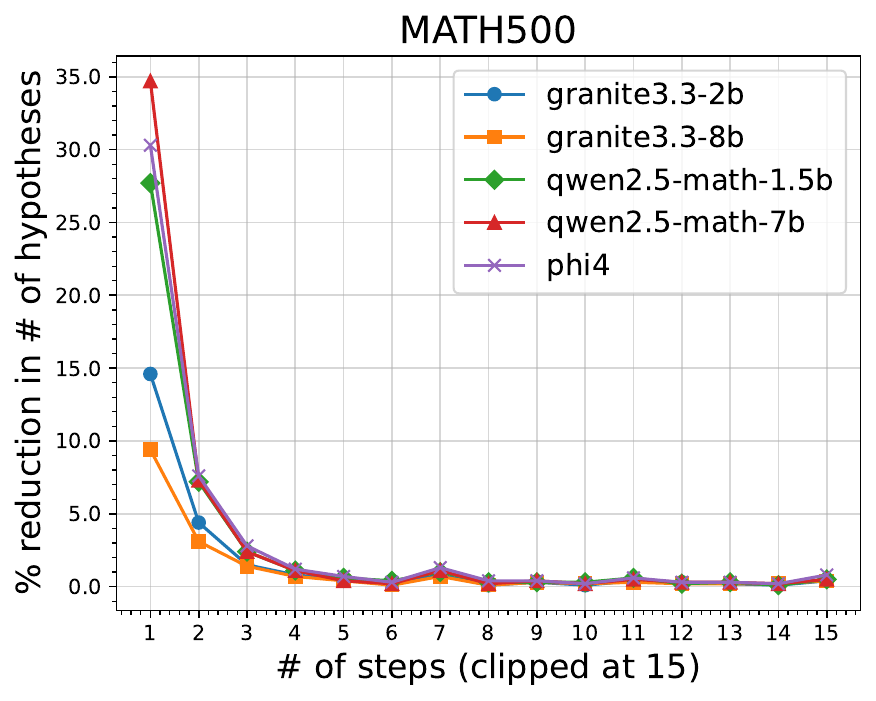}
  \end{subfigure}
  \caption{$\%$ reduction in number of hypotheses as generation progresses for \mathfh{} (initial step size = $256$). More hypotheses are expectedly pruned in earlier steps; $\%$ reduction saturates near a small but non-zero value after $\sim$$6$ steps. \textsc{q}wen and \phifour{} generations are pruned more aggressively as individual hypotheses provide greater lexical coverage of the cohort (\S{\ref{section:experiments}}). \aime{} results are quite similar; c.f. Appendix \ref{appendix:indiv-step-hyp-pruning}.}
  \label{figure:cluster-size-reduction-with-steps-math500}
\end{figure}

\textbf{\textit{A step-by-step look.}}
To further examine the inner working of our method, we take a closer look at the outcome of its individual steps with an initial step size of $256$.
First, as Figure $\ref{figure:cluster-size-reduction-with-steps-math500}$ shows for \mathfh{}, the rate of pruning is expectedly higher in earlier steps of inference, falling gradually over time -- as many prunable items have already been removed -- eventually plateauing near a small non-zero value after about $6$ steps.
See also Appendix \ref{appendix:indiv-step-hyp-pruning}.

Second, we observe slightly different outcomes for different models: \textsc{q}wen and \phifour{} hypotheses get pruned more aggressively than those of \textsc{g}ranite, for instance.
To examine why, we inspect the content of the first $256$ tokens generated by all models with $N=64$.
We find that individual hypotheses from \textsc{q}wen and \textsc{p}hi4 provide a higher token coverage of the group of all $64$ hypotheses compared to \textsc{g}ranite.
For instance, while an average \phifour{} solution prefix of $256$ tokens contains $26.5\%$ of all unique tokens in all $64$ solution prefixes combined, the corresponding number for \granitebig{} is only $15.2\%$.
Thus, when a model's individual hypotheses are lexically more unique within their cohort, the algorithm has to naturally retain more of them to ensure full coverage.

\section{Conclusion}
\label{section:conclusion}

We demonstrate that self-consistency can be made more efficient, while retaining its parallelism, using confidence-weighted token set cover of partial hypotheses.
Future work will explore automatic computation of optimal step size schedules for specific model-problem pairs and application of the method to other domains, \eg{}, code generation.

\section*{Limitations}
While we report extensive evaluation and empirical analysis on the important domain of mathematical problem solving, other long-\chot{} reasoning tasks such as code generation should also benefit from our approach, which we did not explore in this short paper.
It is worth noting though that tasks where a problem can have different final answers that are all correct, including code generation, generally provide a less ideal setting for the application of self-consistency.
A limitation of the proposed method is that it expects a step size schedule as input, which could in principle be estimated using validation data from a diverse set of domains.

\bibliography{custom}

\begin{thebibliography}{18}
\providecommand{\natexlab}[1]{#1}

\bibitem[{Abdin et~al.(2024)Abdin, Aneja et~al.}]{abdin2024phi4}
Marah Abdin, Jyoti Aneja, and 1 others. 2024.
\newblock \href {https://arxiv.org/abs/2412.08905} {{Phi-4 Technical Report}}.
\newblock \emph{Preprint}, arXiv:2412.08905.

\bibitem[{Aggarwal et~al.(2023)Aggarwal, Madaan, Yang, and Mausam}]{aggarwal2023lets}
Pranjal Aggarwal, Aman Madaan, Yiming Yang, and Mausam. 2023.
\newblock \href {https://aclanthology.org/2023.emnlp-main.761/} {{Let's Sample Step by Step: Adaptive-Consistency for Efficient Reasoning and Coding with LLMs}}.
\newblock In \emph{EMNLP}.

\bibitem[{Astudillo et~al.(2025)Astudillo, Sultan, Trivedi, El-Kurdi, Naseem, Florian, and Roukos}]{astudillo2025optimal}
Ramón~Fernandez Astudillo, Md~Arafat Sultan, Aashka Trivedi, Yousef El-Kurdi, Tahira Naseem, Radu Florian, and Salim Roukos. 2025.
\newblock \href {https://arxiv.org/abs/2505.17242} {{Optimal Policy Minimum Bayesian Risk}}.
\newblock \emph{Preprint}, arXiv:2505.17242.

\bibitem[{DeepSeek-AI(2025)}]{deepseekai2025deepseekr1}
DeepSeek-AI. 2025.
\newblock \href {https://arxiv.org/abs/2501.12948} {{DeepSeek-R1: Incentivizing Reasoning Capability in LLMs via Reinforcement Learning}}.
\newblock \emph{Preprint}, arXiv:2501.12948.

\bibitem[{Hendrycks et~al.(2021)Hendrycks, Burns, Kadavath, Arora, Basart, Tang, Song, and Steinhardt}]{hendrycks2021measuring}
Dan Hendrycks, Collin Burns, Saurav Kadavath, Akul Arora, Steven Basart, Eric Tang, Dawn Song, and Jacob Steinhardt. 2021.
\newblock \href {https://openreview.net/forum?id=7Bywt2mQsCe} {{Measuring Mathematical Problem Solving With the MATH Dataset}}.
\newblock In \emph{NeurIPS Track on Datasets and Benchmarks (Round 2)}.

\bibitem[{Huang et~al.(2025)Huang, Huang, Leng, Liu, and Huang}]{huang2025efficient}
Chengsong Huang, Langlin Huang, Jixuan Leng, Jiacheng Liu, and Jiaxin Huang. 2025.
\newblock \href {https://arxiv.org/abs/2503.00031} {{Efficient Test-Time Scaling via Self-Calibration}}.
\newblock \emph{Preprint}, arXiv:2503.00031.

\bibitem[{Lau et~al.(2024)Lau, Hu, Diwen, Jizhuo, Ng, and Low}]{lau2024dipper}
Gregory Kang~Ruey Lau, Wenyang Hu, Liu Diwen, Chen Jizhuo, See-Kiong Ng, and Bryan Kian~Hsiang Low. 2024.
\newblock \href {https://openreview.net/forum?id=0wQsCNrlFl} {{Dipper: Diversity in Prompts for Producing Large Language Model Ensembles in Reasoning tasks}}.
\newblock In \emph{NeurIPS Workshop on Foundation Model Interventions}.

\bibitem[{Li et~al.(2024)Li, Yuan, Feng, Pan, Wang, Sun, Wang, and Li}]{li2024escape}
Yiwei Li, Peiwen Yuan, Shaoxiong Feng, Boyuan Pan, Xinglin Wang, Bin Sun, Heda Wang, and Kan Li. 2024.
\newblock \href {https://openreview.net/forum?id=ndR8Ytrzhh} {{Escape Sky-high Cost: Early-stopping Self-Consistency for Multi-step Reasoning}}.
\newblock In \emph{ICLR}.

\bibitem[{Lightman et~al.(2024)Lightman, Kosaraju, Burda, Edwards, Baker, Lee, Leike, Schulman, Sutskever, and Cobbe}]{lightman2023let}
Hunter Lightman, Vineet Kosaraju, Yuri Burda, Harrison Edwards, Bowen Baker, Teddy Lee, Jan Leike, John Schulman, Ilya Sutskever, and Karl Cobbe. 2024.
\newblock \href {https://openreview.net/forum?id=v8L0pN6EOi} {{Let's Verify Step by Step}}.
\newblock In \emph{ICLR}.

\bibitem[{Mistral-AI(2025)}]{mistralai2025magistral}
Mistral-AI. 2025.
\newblock \href {https://arxiv.org/abs/2506.10910} {Magistral}.
\newblock \emph{Preprint}, arXiv:2506.10910.

\bibitem[{Sun et~al.(2024)Sun, Haider, Zhang, Yang, Qiu, Yin, Wang, Bartlett, and Zanette}]{sun2024fast}
Hanshi Sun, Momin Haider, Ruiqi Zhang, Huitao Yang, Jiahao Qiu, Ming Yin, Mengdi Wang, Peter Bartlett, and Andrea Zanette. 2024.
\newblock \href {https://proceedings.neurips.cc/paper_files/paper/2024/hash/3950f6bf5c2eb7435ecf58eaa85cc8c2-Abstract-Conference.html} {{Fast Best-of-N Decoding via Speculative Rejection}}.
\newblock In \emph{NeurIPS}.

\bibitem[{Touvron et~al.(2023)Touvron, Lavril et~al.}]{touvron2023llama}
Hugo Touvron, Thibaut Lavril, and 1 others. 2023.
\newblock \href {https://arxiv.org/abs/2302.13971} {{LLaMA: Open and Efficient Foundation Language Models}}.
\newblock \emph{Preprint}, arXiv:2302.13971.

\bibitem[{Wan et~al.(2025)Wan, Wu, Chen, and Li}]{wan2025reasoning}
Guangya Wan, Yuqi Wu, Jie Chen, and Sheng Li. 2025.
\newblock \href {https://aclanthology.org/2025.naacl-long.184/} {{Reasoning Aware Self-Consistency: Leveraging Reasoning Paths for Efficient {LLM} Sampling}}.
\newblock In \emph{NAACL}.

\bibitem[{Wang et~al.(2024)Wang, Prasad, Stengel-Eskin, and Bansal}]{wang2024soft}
Han Wang, Archiki Prasad, Elias Stengel-Eskin, and Mohit Bansal. 2024.
\newblock \href {https://aclanthology.org/2024.acl-short.28/} {{Soft Self-Consistency Improves Language Models Agents}}.
\newblock In \emph{ACL}.

\bibitem[{Wang et~al.(2025{\natexlab{a}})Wang, Liu, Chen, Light, Chen, Zhang, and Cheng}]{wang2025diversified}
Tianchun Wang, Zichuan Liu, Yuanzhou Chen, Jonathan Light, Haifeng Chen, Xiang Zhang, and Wei Cheng. 2025{\natexlab{a}}.
\newblock \href {https://arxiv.org/abs/2502.11027} {{Diversified Sampling Improves Scaling LLM inference}}.
\newblock \emph{Preprint}, arXiv:2502.11027.

\bibitem[{Wang et~al.(2023)Wang, Wei, Schuurmans, Le, Chi, Narang, Chowdhery, and Zhou}]{wang2023selfconsistency}
Xuezhi Wang, Jason Wei, Dale Schuurmans, Quoc~V Le, Ed~H. Chi, Sharan Narang, Aakanksha Chowdhery, and Denny Zhou. 2023.
\newblock \href {https://openreview.net/forum?id=1PL1NIMMrw} {{Self-Consistency Improves Chain of Thought Reasoning in Language Models}}.
\newblock In \emph{ICLR}.

\bibitem[{Wang et~al.(2025{\natexlab{b}})Wang, Zhang, Huang, Yang, Zhang, Huang, and Wang}]{wang2025samplingefficient}
Yiming Wang, Pei Zhang, Siyuan Huang, Baosong Yang, Zhuosheng Zhang, Fei Huang, and Rui Wang. 2025{\natexlab{b}}.
\newblock \href {https://arxiv.org/abs/2503.01422} {{Sampling-Efficient Test-Time Scaling: Self-Estimating the Best-of-N Sampling in Early Decoding}}.
\newblock \emph{Preprint}, arXiv:2503.01422.

\bibitem[{Yang et~al.(2024)Yang, Zhang et~al.}]{yang2024qwen2.5}
An~Yang, Beichen Zhang, and 1 others. 2024.
\newblock \href {https://arxiv.org/abs/2409.12122} {{Qwen2.5-Math Technical Report: Toward Mathematical Expert Model via Self-Improvement}}.
\newblock \emph{arXiv preprint arXiv:2409.12122}.

\end{thebibliography}

\appendix

\section{Response Length Statistics}
\label{appendix:token-counts}
Table \ref{table:token-counts} shows the average response length (in number of tokens; without hypothesis pruning) of all \llm{}s we evaluate on the two test sets.
\qwenbig{} is the most verbose of the five models while \qwensmall{} produces the most succinct responses.
Interestingly, we see no clear correlation between response length and performance (reported in \S{\ref{section:experiments}}) for these models.

\begin{table}[h]
    \centering
    \small
    \begin{tabular}{c|cc}
         \textbf{Model} & \textbf{\mathfh{}} & \textbf{\aime{}} \\
         \Xhline{1.6\arrayrulewidth}
         \qwensmall{} & $551.9$ & $867.9$ \\
         \qwenbig{} & $1104.6$ & $2925.7$ \\
         \granitesmall{} & $724.7$ & $1463.6$ \\
         \granitebig{} & $763.8$ & $1502.2$ \\
         \phifour{} & $733.3$ & $1510.1$
    \end{tabular}
    \caption{Average response length ($\#$ of tokens) of different \llm{}s measured in two test sets.}
    \label{table:token-counts}
\end{table}

\begin{table}[t]
    \centering
    \small
    \begin{tabular}{cccc}
        \textbf{Method 1} & \textbf{Method 2} & \textbf{Winner} & \textbf{Win $\%$} \\
        \Xhline{1.6\arrayrulewidth}
        \textsc{cwsc} & \textsc{sc} & \textsc{cwsc} & $82.5$ \\
        \textsc{cwsc} & \textsc{cw} & \textsc{cwsc} & $65.0$ \\
        \textsc{sc} & \textsc{cw} & \textsc{cw} & $62.5$ \\
        \textsc{cwsc} & Random & \textsc{cwsc} & $92.5$ \\
        \Xhline{1.6\arrayrulewidth}
    \end{tabular}
    \caption{Extended ablation results. \textsc{cwsc}: confidence-weighted set cover (proposed); \textsc{sc}: unweighted set cover; \textsc{cw}: confidence weighting with no set cover; Random: random hypothesis pruning. The proposed method outperforms all the simpler methods.}
    \label{table:extended-ablation}
\end{table}

\begin{figure*}[h]
  \centering
  \begin{subfigure}{.4\linewidth}
    \centering
    \includegraphics[width=\linewidth]{figures/math500-pruning-w-steps.pdf}
  \end{subfigure}
  \hspace{1em}%
  \begin{subfigure}{.4\linewidth}
    \centering
    \includegraphics[width=\linewidth]{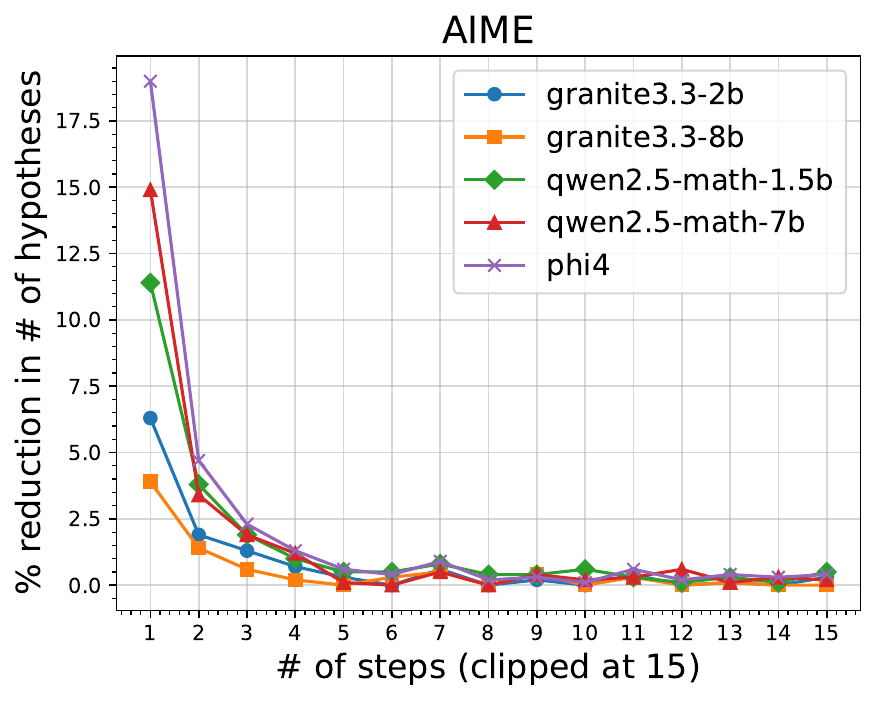}
  \end{subfigure}%
  \hspace{-1em}%
  \caption{$\%$ reduction in number of hypotheses as generation progresses (initial step size = $256$). More hypotheses are pruned in earlier steps of generation; $\%$ reduction saturates near a small but non-zero value after about $5$ or $6$ steps. \textsc{q}wen and \phifour{} generations are pruned more aggressively as individual hypotheses provide greater lexical coverage of the cohort (\S{\ref{section:experiments}}).}
  \label{figure:cluster-size-reduction-with-steps-full}
\end{figure*}

\section{Comparison with Random Pruning}
\label{appendix:compare-with-random}
In Table \ref{table:ablation}, we presented results of ablating the two components of the algorithm -- confidence weighting and token set cover -- separately.
Here we ablate both together by substituting the set $keep$ in line $\ref{algorithm2:call-wsc}$ of Algorithm $\ref{algorithm:efficient-self-consistency}$ with an equal number of randomly chosen hypotheses.
This is in essence also a random baseline that, like \textit{confidence weighting only}, borrows the exact number of hypotheses to prune from the full method, for which it does not have a mechanism of its own.
As the last row of Table \ref{table:extended-ablation} shows, this random baseline is outperformed by the full method in $92.5\%$ of all $40$ evaluations described in \S{\ref{section:experiments}}.
We also show the original ablation results in this table for reference.


\section{Hypothesis Pruning at Different Steps}
\label{appendix:indiv-step-hyp-pruning}

Figure \ref{figure:cluster-size-reduction-with-steps-full} depicts the average $\%$ reduction in number of hypotheses during step-by-step generation of solutions for both \mathfh{} and \aime{}.
A similar trend can be seen on both test sets: a larger fraction of hypotheses get pruned in early stages and fewer in later stages of generation.
On both test sets, \textsc{q}wen and \phifour{} solutions are pruned more aggressively than \textsc{g}ranite solutions; see \S{\ref{section:experiments}} for a detailed discussion on this.

\section{Overall Token Savings}
\label{appendix:token-savings}

Table \ref{table:pct-token-savings} summarizes token savings from the proposed method for all models on both test sets.
The same information is presented in Figure \ref{figure:token-savings-with-pruning} using the symbol $\Delta$ and the two parallel vertical lines corresponding to each short polyline.

\begin{table*}[t]
    \centering
    \small
    \begin{tabular}{clcccc}
        \Xhline{1.6\arrayrulewidth}
        \multirow{3}{*}{\textbf{Benchmark}} & \multicolumn{1}{c}{\multirow{3}{*}{\textbf{Model}}} & \multicolumn{4}{c}{\textbf{$\%$ token savings}} \\
        \cline{3-6}
        & & \multicolumn{4}{c}{\textit{sample budget}} \\
        & & $\mathbf{8}$ & $\mathbf{16}$ & $\mathbf{32}$ & $\mathbf{64}$\\
        \Xhline{1.6\arrayrulewidth}
        \multirow{5}{*}{\mathfh{}} & \qwensmall{} & $14.5$ & $26.8$ & $34.2$ & $33.4$ \\
        & \qwenbig{} & $11.2$ & $15.9$ & $22.8$ & $18.6$ \\
        & \granitesmall{} & $3.3$  & $3.6$ & $15.7$ & $20.1$ \\
        & \granitebig{} & $1.8$ & $7.6$ & $0.4$ & $20.6$ \\
        & \phifour{} & $2.7$ & $1.0$ & $19.5$ & $19.2$ \\
        \hline
        \multirow{5}{*}{\aime{}} & \qwensmall{} & $2.5$ & $0.6$ & $3.0$ & $4.0$ \\
        & \qwenbig{} & $10.2$ & $9.8$ & $13.5$ & $28.3$ \\
        & \granitesmall{} & $4.6$ & $7.3$ & $3.8$ & $7.0$ \\
        & \granitebig{} & $0.1$ & $5.1$ & $2.5$ & $7.1$ \\
        & \phifour{} & $9.4$ & $4.4$ & $12.6$ & $26.8$ \\
        \Xhline{1.6\arrayrulewidth}
    \end{tabular}
    \caption{$\%$ token savings from the proposed confidence-weighted token set cover method for all models and test sets. This is a tabular summary of all $\Delta$ values in Figure \ref{figure:token-savings-with-pruning}. We observe substantial gains of over $10\%$, reaching as high as $34\%+$, in many cases}
    \label{table:pct-token-savings}
\end{table*}

\section{Compute Environment}
\label{appendix:compute-env}
We ran inference with all models on a single \textsc{a}100-80\textsc{gb} \textsc{gpu}; models were served using \textsc{vllm}\footnote{\url{https://github.com/vllm-project/vllm}}.

\end{document}